\documentclass[10pt,twocolumn,letterpaper]{article}

\usepackage[pagenumbers]{cvpr} 

\usepackage[accsupp]{axessibility}

\usepackage{algorithm}
\usepackage{algpseudocode}
\usepackage{amsmath}
\usepackage{booktabs}
\usepackage{multirow}
\usepackage{xcolor,colortbl}

\usepackage[symbol]{footmisc}
\usepackage{lineno}
\definecolor{commentcolor}{RGB}{110,154,155}   

\def\paperID{*****} 
\def\confName{CVPR}
\def\confYear{2025}

\title{Structuring GUI Elements through Vision Language Models: Towards Action Space Generation}

\author{
Yi Xu\textsuperscript{1},
Yesheng Zhang\textsuperscript{1},
Jiajia Liu\textsuperscript{2},
Jingdong Chen\textsuperscript{2}
\\
\textsuperscript{1}Shanghai Jiao Tong University \\
\textsuperscript{2}Ant Group \\
{\tt\small \{yixu98, preacher\}@sjtu.edu.cn}, 
{\tt\small \{lekun.ljj, jingdongchen.cjd\}@antgroup.com}
}

\begin{document}
\maketitle
\begin{abstract}
Multimodal large language models (MLLMs) have emerged as pivotal tools in enhancing human-computer interaction. In this paper we focus on the application of MLLMs in the field of graphical user interface (GUI) elements structuring, where they assist in processing user instructions based on screen contents. Despite the promise of MLLMs, their performance in precisely generating UI element coordinates, a critical aspect of GUI understanding, is hindered by the nature of next-token prediction training. This challenge arises from the semantic void surrounding numerical UI coordinates in language representation spaces, necessitating a substantial and diverse dataset to bolster visual module capabilities. To address these limitations, we introduce an IoU-Augmented Maximum Likelihood (IAML) training paradigm. Specifically, our approach involves a novel pipeline for IoU-based coordinate sampling to augment the training data, which considers the proximity to ground truth coordinates. This data augmentation strategy is then employed to fine-tune MLLMs under the IAML paradigm, which is designed to mitigate the exposure bias problem inherent in traditional maximum likelihood estimation. Through extensive experiments, we demonstrate the superior performance of our IAML training approach over traditional training paradigms.
\end{abstract}    
\section{Introduction}
\label{sec:intro}


Multimodal large language models (MLLMs) have gained great attention in recent years due to their ability to comprehend natural language, visual, and auditory signals, thereby enhancing the productivity of human tasks~\cite{bai2023qwen,wang2024qwen2,yin2023survey,liu2024visual}. Within the realm of graphical user interface (GUI) understanding, MLLMs serve as agents that assist users in executing daily instructions based on screen content~\cite{zhouwebarena,baechler2024screenai,you2024ferret,hong2024cogagent,wang2024mobile,cheng2024seeclick,wu2024atlas,gou2024navigating}, such as clicking specific applications on mobile device or executing commands in desktop or web browsers. Current research endeavors to curate GUI-based datasets for the purpose of instructing MLLMs, endowing them with the capability to ground and refer UI elements. For example, Ferret-UI~\cite{you2024ferret} is equipped with referring, grounding, and reasoning capabilities to recognize and understand screen contents. ScreenAI~\cite{baechler2024screenai} proposes a series of pretraining and fine-tuning mixtures with a wide spectrum of tasks to better tackle UI and infographic challenges.

\begin{figure*}[!t]
    \centering
    \includegraphics[width=1.0\linewidth]{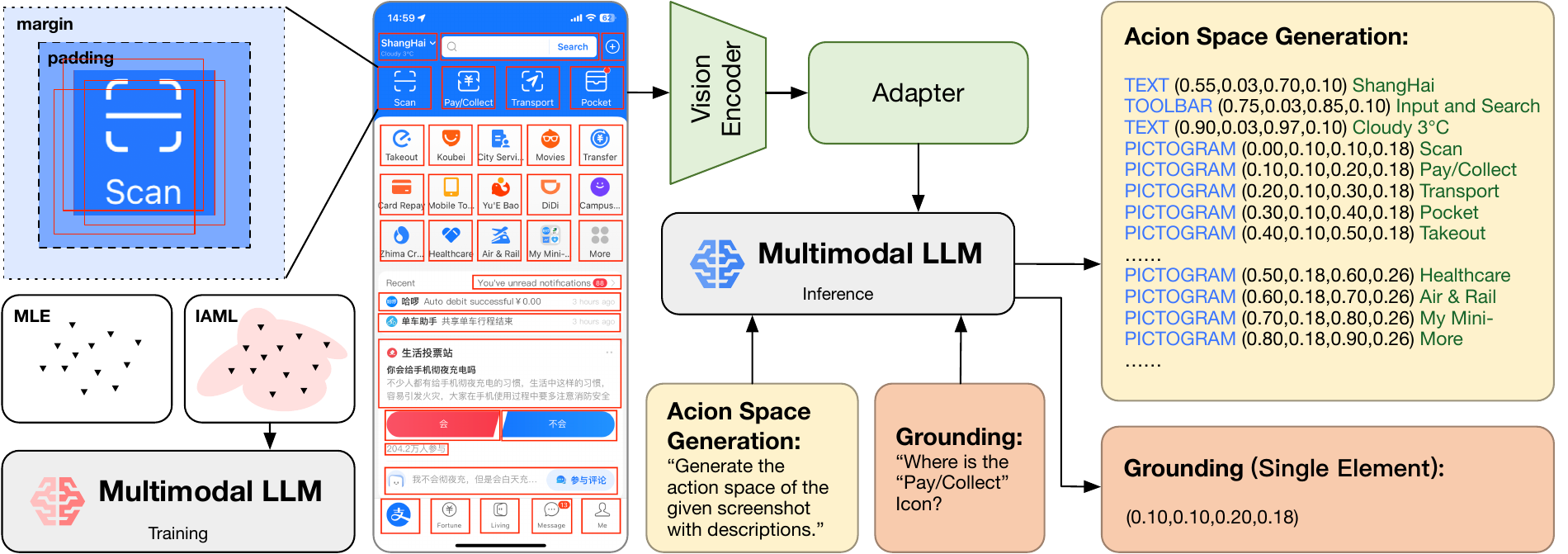}
    \caption{Overview of MLLM agents for GUI grounding tasks and IAML training.}
    \label{framework}
\end{figure*}

However, the inherent nature of next-token prediction training poses challenges for MLLMs in accurately generating the coordinates of UI elements~\cite{cheng2024seeclick,hong2024cogagent}. In general, the bounding box coordinates of a UI element are defined by four values:  ($x\_{min}, y\_{min}, x\_{max}, y\_{max}$). These UI coordinates, being purely numerical and devoid of semantic meaning within the linguistic representation space of LLM, require the spatial awareness that is facilitated by visual modules with adequate UI-coordinate pairs. Consequently, a substantial and diverse dataset of UI information is imperative to reduce the semantic-structure mismatch between language-based generation and visual-based grounding. Recently, several studies have constructed UI-specific datasets~\cite{baechler2024screenai,wu2024mobilevlm,zhang2024android,hong2024cogagent,wu2024atlas,gou2024navigating} through automated multi-step data processes and made them publicly available, yet the quality of these datasets necessitates meticulous consideration. Furthermore, to ground specific UI elements, MLLMs must comprehensively understand the entirety of the screen content, which often encompasses a variety of UI elements constituting a potential action space. To address these challenges, the primary objective of this research is to enhance the capability of MLLMs to recognize comprehensive screen elements and accurately ground coordinates within the framework of auto-regressive generation. Figure~\ref{framework} depicts two kinds of tasks of generating coordinates of UI elements.

For another thing, it is widely acknowledged that decoder-based language models encounter the exposure bias problem~\cite{schmidt2019generalization}, which arises from training with maximum likelihood estimation (MLE) using cross-entropy for next-token prediction. In detail, the model generates the next token based on the ground truth tokens from previous steps in the training process, and during testing, it generates the next word based on the output word rather than ground truth, which makes it difficult to generate precise numerical values. The problem is even worse when dealing with coordinates. Moreover, in the paradigm of cross-entropy training, no matter how far the generated coordinate from the ground truth, once the predictions do not exactly match the ground truth, the MLE treats them equally. Intuitively, we should give more tolerance to those closer predictions than farther ones, which will make the model easier to learn.

Building upon the challenges outlined, we design a new training paradigm that integrates the Intersection-over-Union (IoU) metric with reward-augmented maximum likelihood (RAML)~\cite{norouzi2016reward} for coordinate grounding and action space generation. Our IoU-augmented maximum likelihood is called IAML. It commences with developing an IoU-based coordinate sampling pipeline, which is pivotal for data augmentation. This pipeline meticulously considers the spatial proximity of the sampled coordinates to their ground truth counterparts. Subsequently, the augmented data is fed to fine-tune the MLLM within the MLE training framework. By integrating the IoU-based reward mechanism, we modify the observed distribution (the original training samples) that the MLLM should learn under the MLE framework, making it more accessible and less challenging for the MLLM to grasp. The IAML approach introduces an IoU-based reward system that incentivizes predictions closer to the ground truth, thus alleviating the exposure bias problem inherent in traditional MLE training where all incorrect predictions are penalized equally, regardless of their proximity to the correct answer. The main contributions of this work are summarized as follows:
\begin{itemize}
    \item We analyze the challenges of MLLM for UI element grounding task, which involves the essential issues of language model training and generation.
    \item We propose IoU-Augmented Maximum Likelihood (IAML) training paradigm that bridges the gap between auto-regressive language generation and structured coordinate prediction.
    \item We conduct extensive experiments to validate the performance of our method. The results reveal the efficacy of our method is better than traditional training paradigm and it can be applied to low-resource scenarios. 
\end{itemize}


\section{Related Work}
\subsection{Multimodal Large Language Models (MLLMs)}

Similar to large language models, MLLMs take multimodal data such as images and texts as inputs and auto-regressively generate response token sequence. Let $\mathbf{x}$ denote heterogeneous input tokens including textual and visual tokens, the probability of the response sequence $\mathbf{y}=\{y_i\}^L_{i=1}$ is:

\begin{equation}
p(\mathbf{y}|\mathbf{x}) = \prod_{i=1}^{L}p(y_{i}|\mathbf{x}, {\mathbf{y}_{<i}}).
\end{equation}

During the stage of supervised fine-tuning, MLLMs are trained to maximize the probability $p(\mathbf{y}|\mathbf{x})$ of observed samples from distribution $\mathcal{D}$ by MLE, which is calculated as follows:

\begin{equation}
    \mathcal{L}_{\text{ML}}(\mathcal{D}) = -\sum_{(\mathbf{x},\mathbf{y}) \in \mathcal{D}}\log p(\mathbf{y}|\mathbf{x}).
\end{equation}

Regarding the grounding task, MLLMs engage with numerical coordinate representation in two distinct manners. One approach involves segmenting the image into a grid of approximately 1000 bins and expanding the MLLM's vocabulary to include unique tokens for each bin, thereby encoding coordinates as bin identifiers. Another research direction regards numerical coordinates as a natural language sequence, processing them directly. Our proposed method is versatile, accommodating both strategies for handling coordinates within the MLLM framework.

\subsection{MLLMs as GUI Agents}
With the emergence of Large Language Models (LLMs), LLM-based agents~\cite{xi2023rise,wang2024survey} have demonstrated the ability to tackle sophisticated tasks, thereby assisting humans and alleviating their workload. Researchers have designed various prompting techniques, including in-context learning~\cite{dong2022survey} and reflective strategies~\cite{ji2023towards}, to enhance their efficacy. The advancement of MLLMs has further broadened their capabilities. Some of these vision-based LLMs are now adept at managing real-world demands across various output platforms~\cite{xi2024agentgym,zhouwebarena,baechler2024screenai,you2024ferret,hong2024cogagent,wang2024mobile,cheng2024seeclick,wu2024atlas,gou2024navigating}, such as mobile devices and desktop computers, where the performance of GUI elements grounding is paramount. Recently, MLLM-based agents are capable of handling complex multi-step tasks~\cite{wu2024mobilevlm,zhang2024android,hong2024cogagent}, which need better intra- and inter-UI understanding.

\subsection{Data Smoothing Strategy}
Data smoothing strategies allow larger exploration space surrounding the ground truth samples~\cite{tan2019connectingdotsmlerl}. RAML~\cite{norouzi2016reward} and data noising~\cite{xie2017data,chen2018generative} are two typical methods. RAML means reward augmented maximum likelihood, which is designed to alleviate the exposure bias problem existing in MLE~\cite{schmidt2019generalization}. RAML integrates the concept of reward from reinforcement learning and smooth the sequence-level loss for better generalization. Data noising strategies are more often taken as regularization method, The unigram noising is one of them, which samples nosing data from the unigram frequency distribution to replace each token in the original sequence. Both RAML and data noising strategy expand the exploration space. The augmented sampled data of RAML are based on the task metric reward while those from data noising are based on the unigram frequency.

\section{Method}


\subsection{IAML Overview}
We begin by transforming the classical Reward-Augmented Maximum Likelihood (RAML) formulation into a vision-aware training objective suited for structured coordinate prediction. While RAML was originally proposed for language generation tasks, where rewards are computed using discrete symbolic metrics (e.g., BLEU, edit distance), such metrics fall short in capturing fine-grained spatial differences between bounding boxes in GUI grounding scenarios.

Given an input sequence $\mathbf{x}$ and its corresponding ground truth label $\mathbf{y}$, we define an output hypothesis $\tilde{\mathbf{y}}$ drawn from a candidate space $\tilde{\mathbf{Y}}$. Let $r(\mathbf{y}, \tilde{\mathbf{y}})$ denote the reward of the hypothesis, measured using a task-specific function. In IAML, we instantiate $r(\cdot, \cdot)$ with the Intersection-over-Union (IoU) metric to reflect spatial alignment between bounding boxes. Based on this, we define the exponentiated payoff distribution as:

\begin{equation}
q(\tilde{\mathbf{y}}|\mathbf{y};\tau) = \frac{\exp(r(\mathbf{y}, \tilde{\mathbf{y}}) / \tau)}{\sum_{\tilde{\mathbf{y’}} \in \tilde{\mathbf{Y}}}\exp(r(\mathbf{y}, \tilde{\mathbf{y’}})/\tau)},
\end{equation}
where $\tau$ is a temperature parameter that controls the concentration of the distribution. A smaller $\tau$ emphasizes high-reward outputs, while a larger $\tau$ smooths the distribution for broader exploration.

The IAML training objective follows the RAML framework but redefines both the reward and the augmentation space in a continuous, geometry-aware setting:

\begin{equation}
    \mathcal{L}_{\text{IAML}}(\mathcal{D}) = -\sum_{(\mathbf{x},\mathbf{y}) \in \mathcal{D}}\Big\{\sum_{\tilde{\mathbf{y}} \in \tilde{\mathbf{Y}}}q(\tilde{\mathbf{y}}|\mathbf{y};\tau)\log p(\tilde{\mathbf{y}}|\mathbf{x})\Big\},
\end{equation}
where $p(\tilde{\mathbf{y}}|\mathbf{x})$ denotes the model likelihood under current parameters. Unlike token-level perturbation in language tasks, our method generates augmented outputs by spatially perturbing coordinates around the ground truth, and selecting them based on their IoU scores—effectively constructing a vision-aligned training signal.

In the following section, we present our IoU-based data smoothing strategy, which approximates the exponentiated payoff distribution via Monte Carlo sampling. Then, we detail how IAML enables stable and effective training for both single-element grounding and action space generation tasks in MLLMs.

\subsection{IoU-based Data Smoothing}


%
The original RAML is applied to text generation, which employs hamming or edit distance to construct a reward function. Some researches~\cite{xu2023unsupervised} utilize BLEU metric~\cite{papineni2002bleu} and F1 value to define the reward distribution for structured graph generation. However, it is difficult for text-oriented metrics to deal with the coordinates of bounding boxes since the coordinates are structured. For example, if one of the ground truth coordinate values is $0.88$, text metrics will calculate the tokenized $0.58$ and $0.85$ to be the same distance from the original value, but the latter $0.85$ is closer to $0.88$, which should be allocated higher reward. To this end, we should introduce a vision-based reward metric for GUI tasks.

\begin{algorithm}
\caption{IoU-based Coordinate Sampling}
\label{algo_iou}
\begin{algorithmic}[1]
\Procedure{Augment}{$BBox_{\text{original}}, \tau, \epsilon, N$}
    \State $\mathcal{B} \gets \{\}$

    \For{$i \gets 1$ to $N$}
        \State $\Delta \sim \mathcal{U}(-\epsilon, \epsilon)$
        \State $BBox_{\text{new}} \gets (BBox_{\text{original}} + \Delta)$
        \State $IoU \gets \text{calculate\_iou}(BBox_{\text{original}}, BBox_{\text{new}})$
        \State $I_r \gets \max(0, 99 - \lfloor 100 \times IoU \rfloor)$
        \If{$I_r \notin \mathbb{B}$}
            \State $\mathcal{B}[I_r] \gets \{\}$
        \EndIf
        \State $\mathcal{B}[I_r] \gets \mathcal{B}[I_r] \cup \{BBox_{\text{new}}\}$
    \EndFor

    \State $\mathcal{B} \gets \text{sort}(\mathcal{B})$, $//$ ascending by key
    \State $\mathcal{Q} \gets \mathbf{0}^T$
    \State $i \gets 0$
    \For{$I_{r} \in \mathcal{B}$}
        \State $\mathcal{Q}_i \gets \log(|\mathcal{B}[I_{r}]|) - \frac{I_r}{\tau}$
        \State $i \gets i + 1$
    \EndFor

    \State $\mathcal{Q} \gets Softmax(\mathcal{Q})$

    \State $I_{r'} \gets \text{Random.choice}(\text{len}(\mathcal{Q}), p=\mathcal{Q})$
    \State $BBox_{\text{new}} \gets \text{Random.choice}(\mathcal{B}[I_{r'}], p=\mathcal{U})$
    \State \Return $BBox_{\text{new}}$
\EndProcedure
\end{algorithmic}
\end{algorithm}

The Intersection-over-Union (IoU) is a common evaluation metric used across various fields of computer vision, particularly in object detection tasks, as it quantifies the overlap between two bounding boxes: $IoU = \frac{\text{Area of Overlap}}{\text{Area of Union}}$. In our approach, we employ IoU-based data augmentation to enrich the training data and enhance the model's accuracy in recognizing UI elements. We begin by defining a range for perturbing the bounding box coordinates, which are represented as $(x_{min}, y_{min}, x_{max}, y_{max})$. The coordinate perturbation follows a uniform distribution $(\Delta \sim \mathcal{U}(-\epsilon, \epsilon))$, where $\epsilon$ is the maximum deviation from the original coordinates. It is worth noting that the range of $\epsilon$ is usually limited to the range of the UI element padding length. For example, the augmented bounding boxes of pictogram \textit{Scan} in Figure~\ref{framework} are all in the padding area. This perturbation is applied to the current coordinate of the bounding box, resulting in a new set of coordinates $\mathcal{B}$. One of a coordinate is calculated like:

\begin{equation}
     BBox_{\text{new}} = (x_{\min} + \Delta_x, y_{\min} + \Delta_y, x_{\max} + \Delta_{x'}, y_{\max} + \Delta_{y'}),
\end{equation}
where $\Delta_x, \Delta_y, \Delta_{x'}, \Delta_{y'}$ are independent samples from $\mathcal{U}(-\epsilon, \epsilon)$.

Since the exponential reward distribution of IoU is intractable, we utilize a Monte Carlo method to simulate and approximate the IoU distribution $q(\tilde{\mathbf{y}}|\mathbf{y};\tau)$, allowing us to sample coordinates based on their IoU scores with a higher probability for those with larger IoU scores. Specifically, We perform a Monte Carlo simulation with $N$ iterations. For each iteration, we generate a new bounding box $BBox_{\text{new}}$ and calculate its IoU score according to the original bounding box $BBox_{\text{original}}$. The IoU score is then discretized into the range of $0 \sim 99$ as a reversed index $I_r$:
\begin{equation}
    I_r = \max(0, 99 - \lfloor 100 \times IoU \rfloor).
\end{equation}
This mapping transforms the continuous IoU scores into discrete indices, with lower indices $I_r$ corresponding to higher IoU scores. 


We then calculate the reward distribution based on the index $I_r$, scaling with the number of bounding boxes that map to that index with softmax normalization:

\begin{equation}
    \mathcal{Q} = softmax(\log(|\mathcal{B}[I_{r}]|) - \frac{I_r}{\tau}),
\end{equation}
where $\mathcal{B}[I_{r}]$ is the set of bounding boxes with reward index $I_r$, and $\tau$ is the temperature parameter mentioned above. The term $\log(|\mathcal{B}[I_{r}]|)$ captures the logarithmic count of bounding boxes in each bin associated with the reward index $I_r$. It stabilizes the calculation and converts counts into a comparable scale. $-\frac{I_r}{\tau}$ is proportional to the IoU value, ensuring that higher IoU values (lower $I_r$) receive higher rewards, which aligns with the objective of guiding the model towards generating higher quality bounding boxes. This step effectively balances the model's focus on high-performing predictions with the need to explore a broader range of possibilities, enhancing both the model's ability to generalize. Finally, we sample the new bounding box $BBox_{\text{new}}$ based on the normalized exponentiated payoff distribution $\mathcal{Q}$:

\begin{equation}
    I_{r'} \sim \mathcal{Q},
\end{equation}

\begin{equation}
    BBox_{\text{new}} \sim \mathcal{U}(\mathcal{B}[I_{r'}]),
\end{equation}
where $I_{r'}$ is the index selected according to the distribution $\mathcal{Q}$ and $BBox_{new}$ is uniformly sampled from the set $\mathcal{B}[I_{r'}]$ for bounding boxes associated with $I_{r'}$. This sampling process enriches the training data by favoring bounding boxes with higher IoU scores, thereby improving the model's ability to accurately detect UI elements. The algorithm of IoU-based data sampling are provided in Algorithm~\ref{algo_iou}. 



\begin{figure*}
  \centering
  \begin{subfigure}{0.33\linewidth}
    \includegraphics[width=\linewidth]{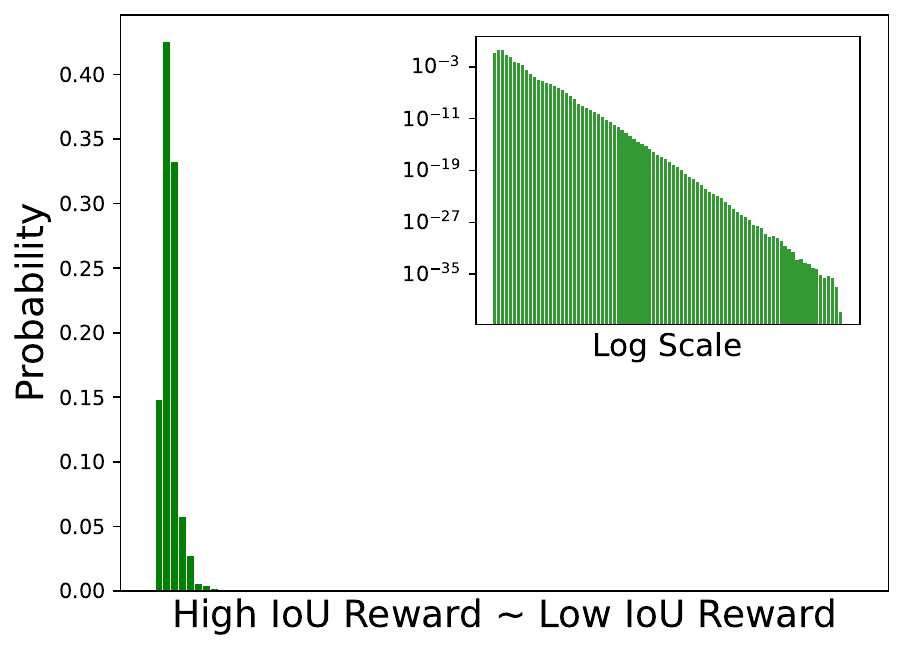}
    \caption{$\tau=1$}
    \label{fig:short-a}
  \end{subfigure}
  \hfill
  \begin{subfigure}{0.33\linewidth}
    \includegraphics[width=\linewidth]{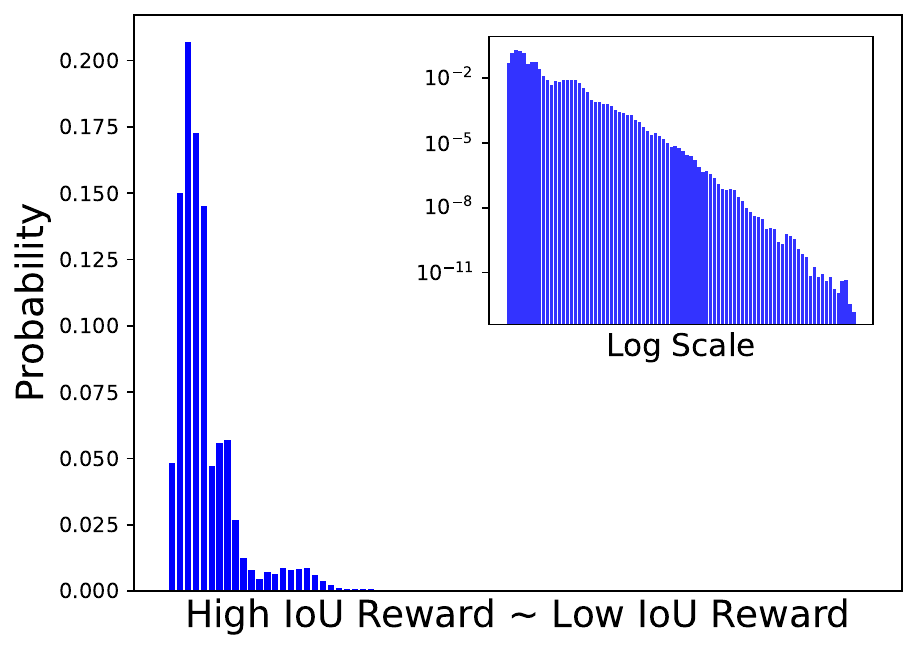}
    \caption{$\tau=3$}
    \label{fig:short-b}
  \end{subfigure}
  \hfill
  \begin{subfigure}{0.33\linewidth}
    \includegraphics[width=\linewidth]{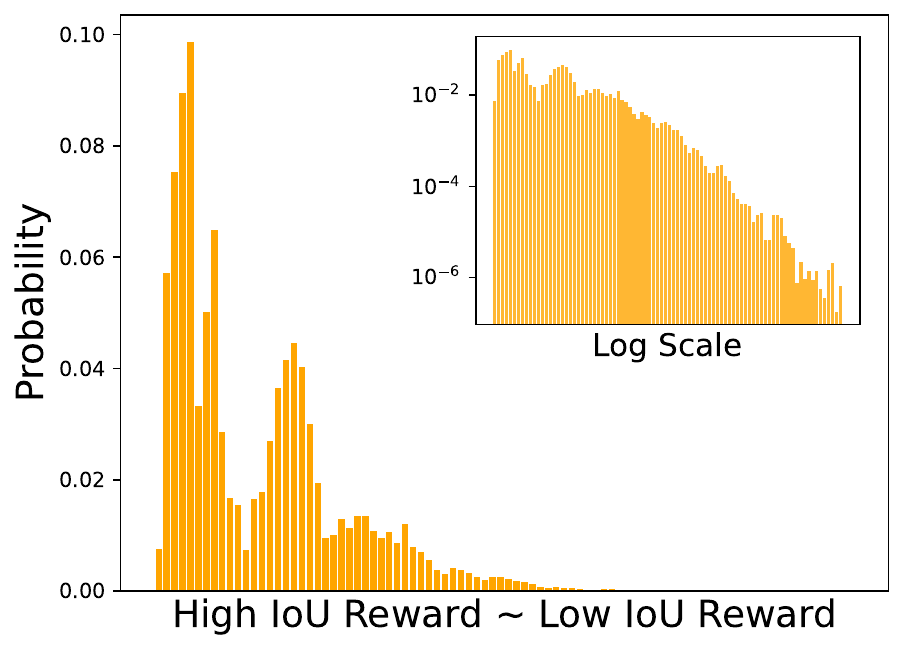}
    \caption{$\tau=6$}
    \label{fig:short-c}
  \end{subfigure}
  \caption{Exponentiated payoff distribution with different $\tau$.}
  \label{fig_dist_tau}
\end{figure*}

\subsection{Model Training and Optimization}
For single element grounding task, our model training incorporates IAML without altering the fundamental MLE loss function. Instead, our training process harnesses the strengths of MLE by maximizing the likelihood of the data sampled from the IoU-based exponentiated payoff distribution $q(\tilde{\mathbf{y}}|\mathbf{y};\tau)$. By augmenting multiple instances and keeping the original ones in the training set, we enable the model to learn from a diverse set of examples.


\begin{table}[!t]
\centering
\tabcolsep 0.17in
\begin{tabular}{c|ccc}
\toprule
                     & \textbf{Mobile} & \textbf{Desktop} & \textbf{Web} \\ \midrule
\textbf{Text}        & 273             & 194              & 230          \\
\textbf{Icon/Widget} & 229             & 140              & 206          \\
\textbf{Total}       & 502             & 334              & 436          \\ \bottomrule
\end{tabular}
\caption{Statistics of ScreenSpot benchmark for inference only.}
\label{table_stats_sa}
\end{table}

\begin{table}[!t]
\centering
\tabcolsep 0.19in
\begin{tabular}{c|ccc}
\toprule
      & \textbf{Training} & \textbf{Validation} & \textbf{Test} \\ \midrule
\textbf{Total} & 15,743    & 2,364       & 4,310 \\ \bottomrule
\end{tabular}
\caption{Statistics of ScreenAnnotation Dataset.}
\label{table_stats_sp}
\end{table}

In the context of action space generation, where a sample may contain multiple elements, our approach involves augmenting each element independently and then concatenating them into a sequence. This method not only creates a rich sequence that captures a broader range of possible actions but also improves the ability of MLLM to handle complex, multi-element scenarios. In detail, the sequence, represented as $\tilde{\mathbf{y}} = (\tilde{\mathbf{y_1}}; \tilde{\mathbf{y_2}}; \ldots; \tilde{\mathbf{y_L}})$, consists of individual hypotheses $\tilde{\mathbf{y_i}}$ for each element within the sample. The model distribution, representing the joint probability distribution of the output $\tilde{\mathbf{y}}$ given the input $\mathbf{x}$, is expressed as:

\begin{equation}
    p(\tilde{\mathbf{y}}|\mathbf{x}) = \prod_{i=1}^{L} p(\tilde{\mathbf{y_i}}|\mathbf{x}, \tilde{\mathbf{y}_{<i}}).
\end{equation}

To measure the difference between the target distribution \(q(\tilde{\mathbf{y}}|\mathbf{y};\tau)\) and the model distribution \(p(\tilde{\mathbf{y}}|\mathbf{x})\), we use the KL-divergence: $D_{KL}(q(\tilde{\mathbf{y}}|\mathbf{y};\tau) \| p_\theta(\tilde{\mathbf{y}}|\mathbf{x}))$ Expanding this divergence, we can decompose it into the sum of KL-divergences for each component $\tilde{\mathbf{y_i}}$:
\begin{equation}
\begin{aligned}
        & \quad\  D_{KL}(q(\tilde{\mathbf{y}}|\mathbf{y};\tau) \| p_\theta(\tilde{\mathbf{y}}|\mathbf{x})) \\
        &= \sum_{\tilde{\mathbf{y}} \in \tilde{\mathbf{Y}}} \prod_{i=1}^{L} q(\tilde{\mathbf{y_i}}|\mathbf{y_i};\tau) \log \frac{\prod_{i=1}^{L} q(\tilde{\mathbf{y_i}}|\mathbf{y_i};\tau)}{\prod_{i=1}^{L} p(\tilde{\mathbf{y_i}}|\mathbf{x}, \tilde{\mathbf{y}_{<i}})} \\
        &= \sum_{\tilde{\mathbf{y}} \in \tilde{\mathbf{Y}}} \prod_{i=1}^{L} q(\tilde{\mathbf{y_i}}|\mathbf{y_i};\tau) \left( \sum_{i=1}^{L} \log \frac{q(\tilde{\mathbf{y_i}}|\mathbf{y_i};\tau)}{p(\tilde{\mathbf{y_i}}|\mathbf{x}, \tilde{\mathbf{y}_{<i}})} \right) \\
        &= \sum_{i=1}^{L} \sum_{\tilde{\mathbf{y}} \in \tilde{\mathbf{Y}}} \prod_{i=1}^{L} q(\tilde{\mathbf{y_i}}|\mathbf{y_i};\tau) \log \frac{q(\tilde{\mathbf{y_i}}|\mathbf{y_i};\tau)}{p(\tilde{\mathbf{y_i}}|\mathbf{x}, \tilde{\mathbf{y}_{<i}})} \\
        &= \sum_{i=1}^{L} ( \sum_{\tilde{\mathbf{y_i}} \in \tilde{\mathbf{Y_i}}} q(\tilde{\mathbf{y_i}}|\mathbf{y_i};\tau) \log \frac{q(\tilde{\mathbf{y_i}}|\mathbf{y_i};\tau)}{p(\tilde{\mathbf{y_i}}|\mathbf{x}, \tilde{\mathbf{y}_{<i}})} ) ( \sum_{j \neq i} q(\tilde{\mathbf{y_j}}|\mathbf{y_j};\tau)) \\
        &= \sum_{i=1}^{L} \sum_{\tilde{\mathbf{y_i}} \in \tilde{\mathbf{Y_i}}} q(\tilde{\mathbf{y_i}}|\mathbf{y_i};\tau) \log \frac{q(\tilde{\mathbf{y_i}}|\mathbf{y_i};\tau)}{p(\tilde{\mathbf{y_i}}|\mathbf{x}, \tilde{\mathbf{y}_{<i}})} \\
        &= \sum_{i=1}^{L} D_{KL}\left(q(\tilde{\mathbf{y_i}}|\mathbf{y_i};\tau) \| p(\tilde{\mathbf{y_i}}|\mathbf{x}, \tilde{\mathbf{y}_{<i}})\right). \\
\end{aligned}
\end{equation}

Minimizing this KL divergence ensures that the model distribution $p(\tilde{\mathbf{y_i}}|\mathbf{x}, \tilde{\mathbf{y}_{<i}})$ converges to the target distribution for each element $q(\tilde{\mathbf{y_i}}|\mathbf{y_i};\tau)$, which is also minimizing the IAML loss since:

\begin{equation}
\begin{aligned}
    & \quad\ \mathcal{L}_{\text{IAML}}((\mathbf{x},\mathbf{y_i})) = -\sum_{\tilde{\mathbf{y}} \in \tilde{\mathbf{Y}}}q(\tilde{\mathbf{y_i}}|\mathbf{y_i};\tau)\log p(\tilde{\mathbf{y_i}}|\mathbf{x}, \tilde{\mathbf{y}_{<i}}) \\
    &= D_{KL}\left(q(\tilde{\mathbf{y_i}}|\mathbf{y_i};\tau) \| p(\tilde{\mathbf{y_i}}|\mathbf{x}, \tilde{\mathbf{y}_{<i}})\right) + \mathbb{H}(q(\tilde{\mathbf{y_i}}|\mathbf{y_i};\tau)),
\end{aligned}
\end{equation}
where $\mathbb{H}(q(\tilde{\mathbf{y_i}}|\mathbf{y_i};\tau))$ is a constant and represents the entropy of the target distribution. Thus, we demonstrate the effectiveness of an output composed of multiple independent samples based on different IoUs in the action space generation problem. This method ensures that the model can learn the correct distribution of each $\tilde{\mathbf {y_i}}$, thereby improving the overall prediction accuracy.

\section{Experiments}
To evaluate the performance and effectiveness of our proposed method, we conduct different kinds of experiments including comparisons on action space generation and single element grounding task. We compare our method with SOTA baselines using MLE training and random data augmentation. Besides, we also analyze hyper-parameters effects during the training stage.


\begin{table*}[htb]
\centering
\tabcolsep 0.04in
\renewcommand{\arraystretch}{1.35}
\begin{tabular}{c|c|ccc|ccc|ccc|ccc}
\toprule
\multirow{2}{*}{\textbf{Data}}     & \multirow{2}{*}{\textbf{Method}} & \multicolumn{3}{c|}{\textbf{IoU=0.1}} & \multicolumn{3}{c|}{\textbf{IoU=0.3}} & \multicolumn{3}{c|}{\textbf{IoU=0.5}} & \multicolumn{3}{c}{\textbf{IoU=0.7}} \\
                                   &                                     & \textbf{P}  & \textbf{R} & \textbf{F} & \textbf{P}  & \textbf{R} & \textbf{F} & \textbf{P}  & \textbf{R} & \textbf{F} & \textbf{P} & \textbf{R} & \textbf{F} \\ \midrule
\multirow{4}{*}{all training set}  & Random                              & 0.799       & 0.547      & 0.630      & 0.744       & 0.512      & 0589       & 0.658       & 0.455      & 0.523      & 0.523      & 0.364      & 0.418      \\
                                   & MLE                                 & 0.794       & 0.541      & 0.626      & 0.736       & 0.503      & 0.582      & 0.652       & 0.448      & 0.517      & 0.526      & 0.364      & 0.419      \\
                                   & IAML                                & 0.803       & 0.548      & 0.632      & 0.747       & 0.512      & 0.591      & 0.662       & 0.457      & 0.526      & 0.538      & 0.373      & 0.429      \\
                            \rowcolor{gray!20}& $\uparrow$                         & 1.10\%      & 1.28\%     & 1.03\%     & 1.49\%      & 1.79\%     & 1.50\%     & 1.67\%      & 2.02\%     & 1.73\%     & 2.17\%     & 2.32\%     & 2.14\%     \\ \midrule
\multirow{3}{*}{10\% training set} & MLE                                 & 0.610       & 0.332      & 0.409      & 0.493       & 0.265      & 0.326      & 0.355       & 0.188      & 0.233      & 0.224      & 0.121      & 0.148      \\
                                   & IAML                                & 0.646       & 0.344      & 0.422      & 0.510       & 0.279      & 0.339      & 0.371       & 0.204      & 0.248      & 0.227      & 0.127      & 0.154      \\
                                   \rowcolor{gray!20}& $\uparrow$                         & 5.90\%      & 3.61\%     & 3.18\%     & 3.45\%      & 5.28\%     & 3.99\%     & 4.51\%      & 8.51\%     & 6.44\%     & 1.34\%     & 4.96\%     & 4.05\%     \\ \bottomrule
\end{tabular}
\caption{Results of SeeClick for action space generation on ScreenAnnotation dataset. $\uparrow$ represents the improvement of IAML relative to MLE.}
\label{table_asg_all}
\end{table*}

\subsection{Datasets}
To evaluate the performance of method for action space generation, we use the ScreenAnnotation dataset~\cite{baechler2024screenai}, which consists of 4,200 mobile screenshots from the Rico dataset~\cite{deka2017rico}. We compose the elements information as the format $(\textit{type}, \textit{coordinates}, \textit{description})$. For single element GUI grounding, we use the ScreenSpot dataset~\cite{cheng2024seeclick}, which containing different types of elements from mobile, desktop and web contents. For the convenience of bounding box process, we map the original coordinates between 0 and 1. The statistics of these two datasets are listed in Table~\ref{table_stats_sa} and Table~\ref{table_stats_sp}.

\subsection{Experimental Settings}
We use existing MLLMs including SeeClick~\cite{cheng2024seeclick}, and OS-Atlas~\cite{wu2024atlas} for continual pre-training with our proposed strategy since these two models are tuned with large amounts of GUI datasets. It should be noted that SeeClick model is trained from Qwen-VL-Chat~\cite{bai2023qwen} while OS-Atlas is based on Qwen2-VL-7B-Instruct~\cite{wang2024qwen2}. We use the configurations recommended in paper~\cite{cheng2024seeclick}. The gradients of the visual encoder are unlocked and the text decoder is all fine-tuned with LoRA~\cite{DBLP:conf/iclr/HuSWALWWC22}. To ensure a fair comparison, we conduct $k$ epochs of training using MLE. We sample $k$ instances, including the original sample, and perform one training epoch using the IAML approach. This setup ensures that the total number of training steps remains consistent across both methods and that the position of samples within the two training sets remains constant.  All experiments are conducted on NVIDIA A800 GPUs (80G×8).

\begin{figure}[htb]
    \centering
    \includegraphics[width=0.95\linewidth]{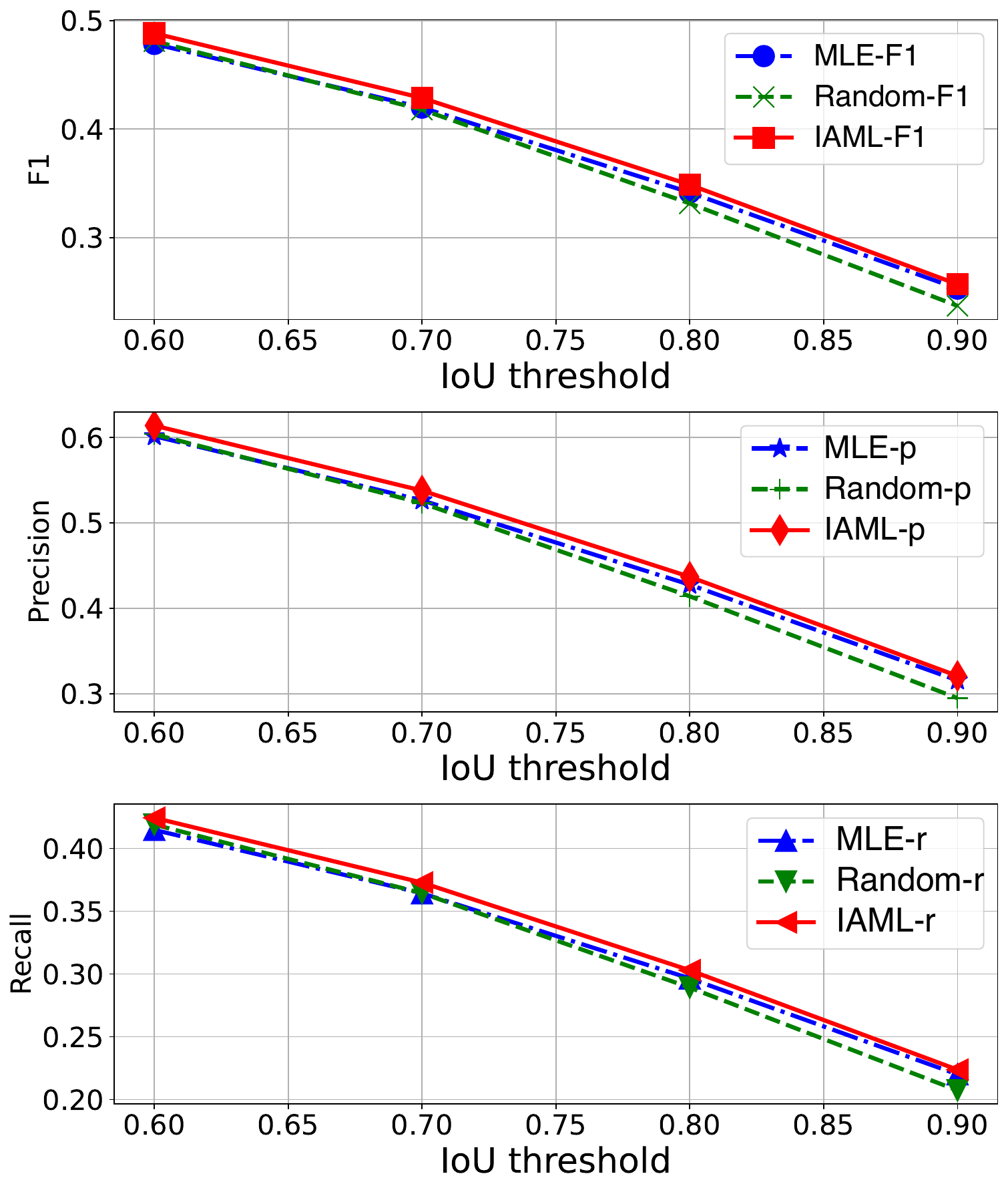}
    \caption{The performance (precision, recall, and F1) of MLE and IAML under different IoU thresholds.}
    \label{mle_raml_iou}
\end{figure}

To evaluate the performance of different methods for action space generation, we use precision ($P$), recall ($R$), and F1 score ($F$) as our primary metrics. These metrics are calculated under different IoU thresholds. A predicted bounding box is deemed a match if its IoU score with the ground truth surpasses the designated threshold. As to single element GUI grounding, we calculate the click accuracy by determining if the agent successfully clicks on the correct position, aligning with the metric utilized in SeeClick~\cite{cheng2024seeclick}.

\begin{table*}[htb]
\centering
\tabcolsep 0.046in
\renewcommand{\arraystretch}{1.35}
\begin{tabular}{c|c|c|ccc|ccc|ccc|c}
\toprule[1pt]
\multirow{2}{*}{\textbf{LLM}} & \multirow{2}{*}{\textbf{Size}} & \multirow{2}{*}{\begin{tabular}[c]{@{}c@{}}\textbf{GUI}\\ \textbf{Tuned}\end{tabular}} & \multicolumn{3}{c|}{\textbf{Mobile}} & \multicolumn{3}{c|}{\textbf{Desktop}} & \multicolumn{3}{c|}{\textbf{Web}} & \multirow{2}{*}{\begin{tabular}[c]{@{}c@{}}\textbf{Average}\\ \textbf{Accuracy}\end{tabular}} \\
                     &                       &                                                                      & \textbf{Text}    & \textbf{Icon}    & \textbf{AVG.}    & \textbf{Text}     & \textbf{Icon}    & \textbf{AVG.}    & \textbf{Text}   & \textbf{Icon}   & \textbf{AVG.}   &                                                                             \\ \midrule
MiniGPT-v2           & 7B                    & $\times$                                                             & 8.4\%   & 6.6\%   & 7.5\%   & 6.2\%    & 2.9\%   & 4.6\%   & 6.5\%  & 3.4\%  & 5.0\%  & 5.7\%                                                                       \\
Qwen-VL              & 9.6B                  & $\times$                                                             & 9.5\%   & 4.8\%   & 7.2\%   & 5.7\%    & 5.0\%   & 5.4\%   & 3.5\%  & 2.4\%  & 3.0\%  & 5.2\%                                                                       \\
GPT-4V               & -                     & $\times$                                                             & 22.6\%  & 24.5\%  & 23.6\%  & 20.2\%   & 11.8\%  & 16.0\%  & 9.2\%  & 8.8\%  & 9.0\%  & 16.2\%                                                                      \\
GPT-4o               & -                     & $\times$                                                             & 20.2\%  & 24.9\%  & 22.6\%  & 21.1\%   & 23.6\%  & 22.4\%  & 12.2\%  & 7.8\%  &10.0\%  & 18.3\%                                                                      \\
Fuyu                 & 18B                   & \checkmark                                            & 41.0\%  & 1.3\%   & 21.2\%  & 33.0\%   & 3.6\%   & 18.3\%  & 33.9\% & 4.4\%  & 19.2\% & 19.5\%                                                                      \\
CogAgent             & 18B                   & \checkmark                                            & 67.0\%  & 24.0\%  & 45.5\%  & 74.2\%   & 20.0\%  & 47.1\%  & 70.4\% & 28.6\% & 49.5\% & 47.4\%                                                                      \\
Qwen2-VL             & 8.3B                   & \checkmark                                            & 61.3\%  & 39.3\%  & 50.3\%  & 52.0\%   & 45.0\%  & 48.5\%  & 33.1\% & 21.8\% & 27.5\% & 42.9\%                                                                      \\ \midrule
SeeClick-Weight      & 9.6B                  & \checkmark                                            & 77.7\%  & 50.7\%  & 64.2\%  & 70.1\%   & 31.4\%  & 50.8\%  & 52.2\% & \textbf{32.5\%} & 42.4\% & 52.4\%                                                                      \\
SeeClick-MLE         & 9.6B                  & \checkmark                                            & 79.1\%  & 52.4\%  & 65.8\%  & 67.0\%   & \textbf{36.4\%}  & 51.7\%  & 52.6\% & 28.6\% & 40.6\% & 52.7\%                                                                      \\ 
\rowcolor{gray!20}SeeClick-IAML        & 9.6B                  & \checkmark                                            & \textbf{80.2\%}  & \textbf{53.3\%}  & \textbf{66.8\%}  & \textbf{71.6\%}   & 34.3\%  & \textbf{53.0\%}  & \textbf{61.3\%} & 29.1\% & \textbf{45.2\%} & \textbf{55.0\%}                                                                      \\ \midrule
OS-Atlas-Weight         & 8.3B                  & \checkmark                                            & 73.6\%  & 48.5\%  & 61.1\%  & 51.5\%   & 21.4\%  & 36.5\%  & 47.4\% & 33.0\% & 40.2\% & 45.9\%                                                                      \\ 
OS-Atlas-MLE         & 8.3B                  & \checkmark                                            & 78.8\%  & 52.8\%  & 65.9\%  & 42.8\%   & 18.6\%  & 30.7\%  & 61.7\% & 41.3\% & 51.5\% & 49.4\%                                                                      \\ 
\rowcolor{gray!20}OS-Atlas-IAML        & 8.3B                  & \checkmark                                            & \textbf{81.3\%}  & \textbf{60.3\%}  & \textbf{70.8\%}  & \textbf{62.9\%}   & \textbf{39.3\%}  & \textbf{51.1\%}  & \textbf{69.6\%} & \textbf{45.6\%} & \textbf{57.6\%} & \textbf{59.8\%}                                                                      \\ \bottomrule

\end{tabular}
\caption{Comparision results on ScreenSpot dataset.}
\label{table_cmp_single}
\end{table*}

\begin{table*}[!tb]
\centering
\tabcolsep 0.03in
\renewcommand{\arraystretch}{1.35}
\begin{tabular}{c|c|ccc|ccc|ccc|ccc}
\toprule
\multirow{2}{*}{\textbf{Model}}         & \multirow{2}{*}{\textbf{Method}} & \multicolumn{3}{c|}{\textbf{IoU=0.1}} & \multicolumn{3}{c|}{\textbf{IoU=0.3}} & \multicolumn{3}{c|}{\textbf{IoU=0.5}} & \multicolumn{3}{c}{\textbf{IoU=0.7}} \\
                          &                                     & \textbf{P}  & \textbf{R} & \textbf{F} & \textbf{P}  & \textbf{R} & \textbf{F} & \textbf{P}  & \textbf{R} & \textbf{F} & \textbf{P} & \textbf{R} & \textbf{F} \\ \midrule
\multirow{3}{*}{Qwen2-VL} & MLE                                 & 0.406       & 0.266      & 0.289      & 0.262       & 0.174      & 0.188      & 0.162       & 0.107      & 0.116      & 0.093      & 0.063      & 0.068      \\
                          & IAML                                & 0.420       & 0.327      & 0.340      & 0.277       & 0.217      & 0.225      & 0.174       & 0.139      & 0.143      & 0.112      & 0.090      & 0.091      \\
                          \rowcolor{gray!20}& $\uparrow$                         & 3.25\%      & 23.05\%    & 17.44\%    & 5.71\%      & 24.84\%    & 19.66\%    & 7.37\%      & 29.94\%    & 23.19\%    & 20.02\%    & 41.41\%    & 35.54\%    \\ \midrule
\multirow{3}{*}{OS-Atlas} & MLE                                 & 0.602       & 0.352      & 0.423      & 0.446       & 0.263      & 0.316      & 0.294       & 0.177      & 0.211      & 0.165      & 0.101      & 0.120      \\
                          & IAML                                & 0.612       & 0.372      & 0.441      & 0.463       & 0.286      & 0.339      & 0.318       & 0.201      & 0.237      & 0.192      & 0.123      & 0.144      \\
                          \rowcolor{gray!20}& $\uparrow$                         & 1.66\%      & 5.57\%     & 4.40\%     & 3.94\%      & 8.67\%     & 7.21\%     & 7.99\%      & 13.69\%    & 11.90\%    & 16.20\%    & 22.08\%    & 20.37\%    \\ \bottomrule
\end{tabular}
\caption{Results of Qwen2-VL and OS-Atlas for action space generation on ScreenAnnotation Dataset.}
\label{table_asg_qwen}
\end{table*}

\subsection{Results of Action Space Generation}

We first test the improvement of our proposed method on action space generation task based on SeeClick, where the challenge lies in predicting multiple screen elements including \textit{type}, \textit{coordinates}, and \textit{description}. Our method was compared with a random data noising strategy and the traditional MLE training loss to assess its efficacy. The comparative analysis is detailed in Table~\ref{table_asg_all}, which reveals that our IAML approach outperforms both the random noising strategy and the MLE training. Specifically, IAML demonstrates a minimum of 1\% relative improvement over MLE, underscoring its superiority in action space generation. While the random data noising strategy also outperforms MLE, it falls short in comparison to our IAML method. We can also observe from the Table~\ref{table_asg_all} and Figure~\ref{mle_raml_iou} that as the IoU threshold increases, the advantage of IAML becomes more apparent, indicating that our method can make the generated coordinates closer to the real coordinates, that is, more accurate.

To further validate the robustness of our approach, especially in scenarios where data is scarce, we conducted additional tests under a low-resource condition, utilizing only $10\%$ of the original dataset. The results are telling: the IAML method exhibits an even higher relative improvement in performance compared to the full dataset setting. This not only substantiates the effectiveness of our method in augmenting GUI data but also highlights its resilience and adaptability in environments with limited training samples. Additionally, we employ Qwen2-VL-7B-Instruct~\cite{wang2024qwen2} and OS-Atlas-Base-7B~\cite{wu2024atlas} as the base models to train the ScreenAnnotation dataset, with the outcomes presented in Table~\ref{table_asg_qwen}. Although the performance of these two models slightly trails that of the SeeClick model, they reliably affirm our earlier observations: the IoU-based IAML method consistently outperforms the MLE approach, where OS-Atlas-IAML achieves a minimum of 4.4\% higher F1 scores compared to the MLE method.

\subsection{Results of Single Element GUI Grounding}
It is crucial to empower MLLM with the ability to understand the action space, however, in the real-world scenario, users sometimes give a single instruction to MLLM to manipulate the screen such as the click operation. Therefore, in this part, we test the performance of IAML method for single element GUI grounding. The accuracy results are shown in Table~\ref{table_cmp_single}. Like the experiments in \cite{cheng2024seeclick}, we select two types of baseline models for comparison: one category consists of models without GUI fine-tuning, including MiniGPT-v2~\cite{chen2023minigpt}, Qwen-VL~\cite{bai2023qwen}, GPT-4V and GPT-4o. The other category comprises models that have undergone GUI fine-tuning, which includes Fuyu~\cite{fuyu-8b}, CogAgent~\cite{hong2024cogagent}, Qwen2-VL~\cite{wang2024qwen2}, and variants based on SeeClick~\cite{cheng2024seeclick}, OS-Atlas~\cite{wu2024atlas} architectures. 

The SeeClick-Weight and OS-Atlas-Weight directly incorporate rewards into the maximum likelihood loss which is shown in Equation~\ref{mle_weight}. The SeeClick-MLE and OS-Atlas-MLE represent the results of further supervised fine-tuning on our data. Notably, the SeeClick-Weight exhibits a marginal decrease in performance, while the OS-Atlas-Weight lags significantly, which can be attributed to the instability in model training caused by directly adding rewards to the training loss, leading to difficulties in loss convergence. In contrast, our proposed method SeeClick-IAML and OS-Atlas-IAML demonstrate superior performance across the vast majority of platforms, achieving the best average performance on mobile devices, computers, and web interfaces, with OS-Atlas-IAML emerging as the most effective.

\begin{equation}\label{mle_weight}
    \mathcal{L}_{\text{weight}}(\mathcal{D}) = -\sum_{(\mathbf{x},\mathbf{y}) \in \mathcal{D}}\Big\{\sum_{\tilde{\mathbf{y}} \in \tilde{\mathbf{Y}}}r(\mathbf{y},\tilde{\mathbf{y}}) \cdot \log p(\tilde{\mathbf{y}}|\mathbf{x})\Big\}.
\end{equation}



\begin{figure}
  \centering
  \begin{subfigure}{0.49\linewidth}
    \includegraphics[width=\linewidth]{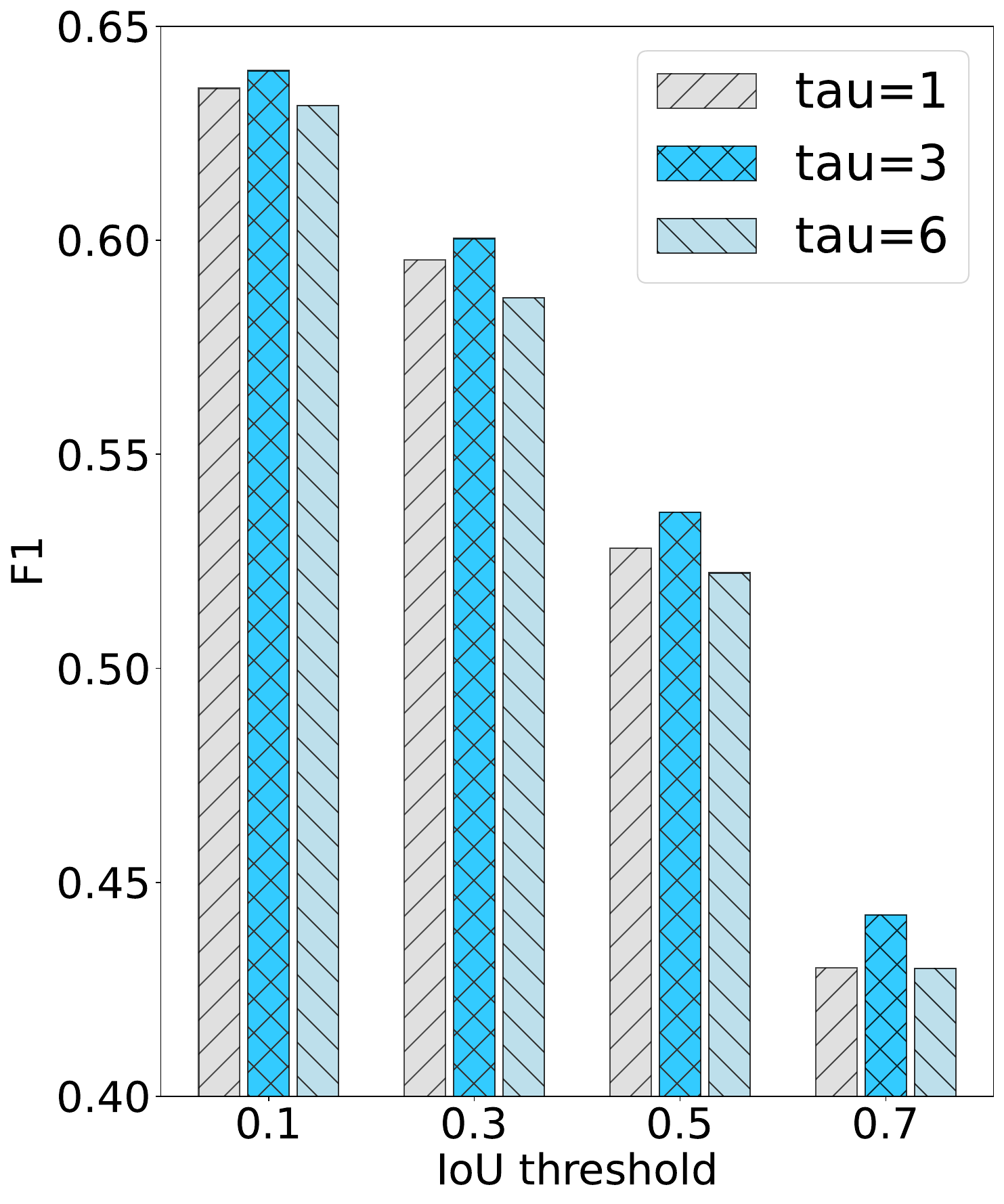}
    \caption{Hyper-parameter $\tau$}
    \label{tau_bar}
  \end{subfigure}
  \hfill
  \begin{subfigure}{0.49\linewidth}
    \includegraphics[width=\linewidth]{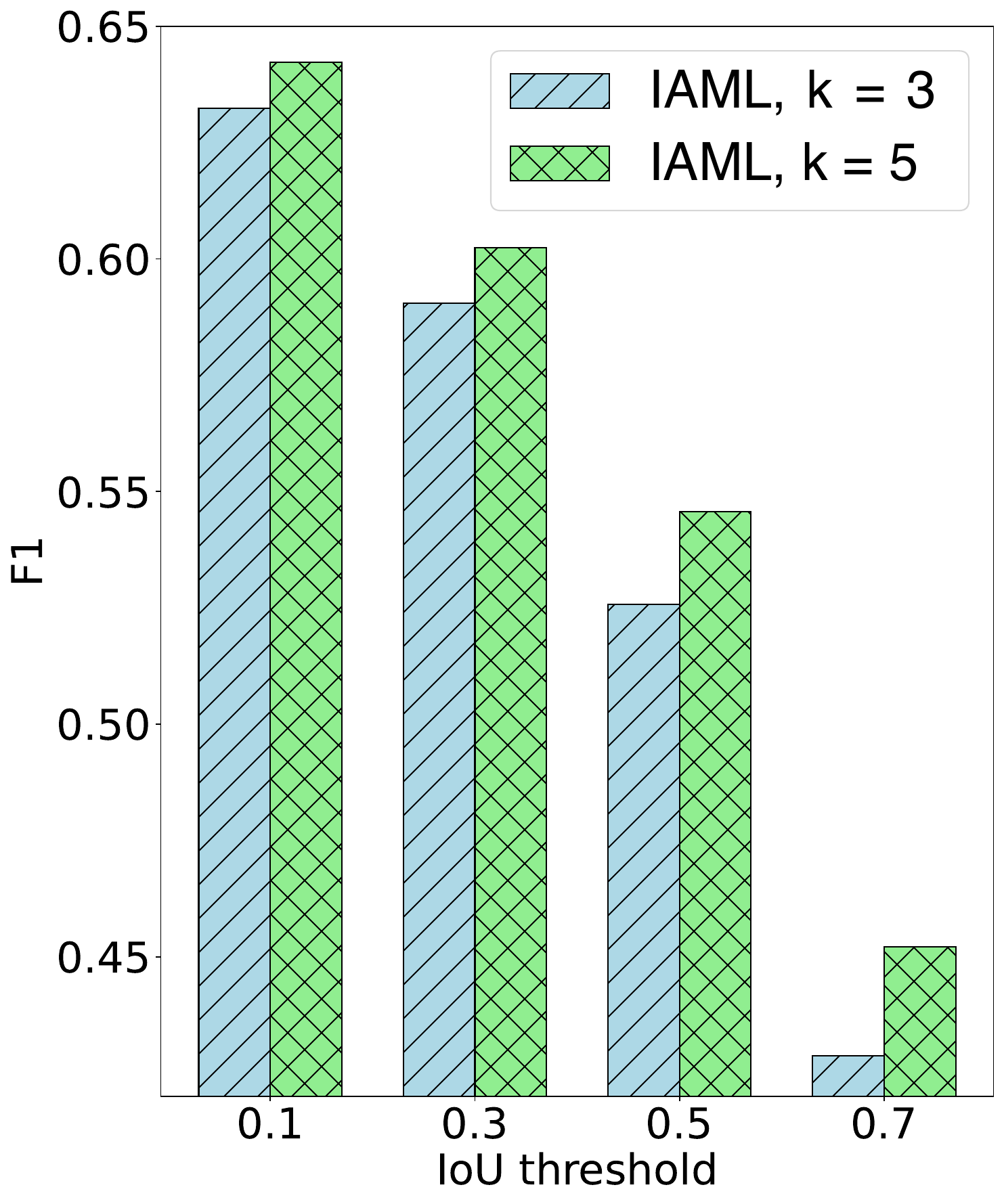}
    \caption{Hyper-parameter $k$}
    \label{samples_bar}
  \end{subfigure}
  \caption{The performance of IAML with different $\tau$ values and numbers of augmented samples $k$. }
  \label{fig_hyper}
\end{figure}

\subsection{Effects of Hyper-parameters}
There are two important hyper-parameters in our method. One is the temperature $\tau$ for reward distribution in IAML loss. The other is the number of sampled instances $k$ (including the original instance). Figure~\ref{tau_bar} delineates the model's performance across varying values of $\tau$, specifically $\tau=1, 3,$ and $6$. It is evident that the optimal performance is achieved when $\tau$ is set to $3$, a distributional profile that is further detailed in Figure~\ref{fig:short-b}. This particular value of $\tau$ strikes an optimal balance, ensuring that the model maintains a keen focus on coordinates that yield high rewards while also encouraging the exploration of a diverse solution space. 

In addition, figure~\ref{samples_bar} substantiates the positive correlation between the number of augmented samples $k$ and the model's performance, indicating that an increased number of samples leads to a steady enhancement in performance. This suggests that data augmentation is a valuable strategy for improving the model's ability to generalize and predict accurately. However, in the practical application of IAML, a delicate balance must be struck between the pursuit of performance and the considerations of training efficiency.

\section{Conclusion}

In this paper, we introduce an IoU-Augmented Maximum Likelihood (IAML) approach tailored for GUI agents to tackle the challenge of precise screen content generation with accurate UI element coordinates. Our method involves augmenting samples based on the IoU reward distribution between the original and perturbed bounding boxes, which facilitates the learning process for the MLLM by expanding its exposure to a broader range of potential outputs. This augmentation strategy not only mitigates the inherent bias in numerical coordinate generation present in MLLMs, stemming from their traditional next-token prediction training, but also enhances the model's ability to accurately generate coordinates. Our experimental results demonstrate that IAML approach significantly boosts performance in both action space generation and single element grounding tasks. Furthermore, the flexibility of our method allows it to be particularly effective in low-resource settings where limited training samples are available. In the future, we will explore the theoretical upper limit of the ability of IAML and verify its performance on ultra long structured GUI elements.

{
    \small
    \bibliographystyle{ieeenat_fullname}
    \bibliography{main}
}


\end{document}